# Optimistic Temporal Difference Learning for 2048

Hung Guei, Lung-Pin Chen, *Member IEEE*, and I-Chen Wu, *Senior Member, IEEE*

*Abstract*—Temporal difference (TD) learning and its variants, such as multistage TD (MS-TD) learning and temporal coherence (TC) learning, have been successfully applied to 2048. These methods rely on the stochasticity of the environment of 2048 for exploration. In this paper, we propose to employ optimistic initialization (OI) to encourage exploration for 2048, and empirically show that the learning quality is significantly improved. This approach optimistically initializes the feature weights to very large values. Since weights tend to be reduced once the states are visited, agents tend to explore those states which are unvisited or visited few times. Our experiments show that both TD and TC learning with OI significantly improve the performance. As a result, the network size required to achieve the same performance is significantly reduced. With additional tunings such as expectimax search, multistage learning, and tile-downgrading technique, our design achieves the state-of-the-art performance, namely an average score of 625 377 and a rate of 72% reaching 32768-tiles. In addition, for sufficiently large tests, 65536-tiles are reached at a rate of 0.02%.

*Index Terms*—2048, $n$-tuple network, optimistic initialization, reinforcement learning, stochastic puzzle games, temporal difference (TD) learning.

## I. Introduction

*2048* is a single-player stochastic puzzle game introduced by Cirulli [1] as a variant of *Threes!* and *1024*. This intriguing and even addictive game has been popular worldwide since it is non-trivial to master despite the simple rules [2], and has also attracted researchers to develop game-playing programs [3]. Due to its simplicity and complexity [4], 2048 is considered to be an interesting and challenging platform for evaluating the effectiveness of *machine learning* methods [5], [6], [7].

In the past, many methods were proposed for 2048. Szubert and Jaśkowski [2] applied the *temporal difference* (TD) *learning* with $n$-tuple networks to 2048. In their approach, 2048-tiles were achieved at a rate of 97%. Yeh *et al.* [3] introduced the *multistage TD* (MS-TD) *learning* which improved the training by separating an entire episode into several stages. Their 3-stage TD method reached 32768-tiles with a rate of 31.75% and even achieved the first-ever 65536-tile. Matsuzaki [8] presented a systematic analysis on $n$-tuples, identified some best configurations of 8×6-tuples and 8×7-tuples. Jaśkowski [5] improved the performance with *temporal coherence* (TC) *learning*, which accelerated the convergence of the training by adaptively reducing the learning rate. They reached state-of-the-art (SOTA) results as follows. The 32768-tiles reaching rate was even improved up to 70%, by using a 16-stage TC method with one second per move.

Although various methods have been continually proposed and improved, these TD methods for 2048, however, were still based on the greedy policy [9] to deterministically choose actions with maximum estimations, and thus relied on the stochastic environment to provide enough randomness for exploration. In addition, it is observed that TD learning tends not to reach large tiles when training is saturated in terms of the average score, namely when the average score increases to nearly the highest [3], [10]. This reflects a potential problem of exploration deficiency. On the other hand, recent works [3], [8], [11] tended to employ larger $n$-tuple networks for higher performance. Hence, as the networks become larger, the issue of exploration deficiency becomes non-negligible. Moreover, in addition to the size of the network, the learning rate is another factor related to exploration deficiency. TC learning effectively improved the performance but converged rather fast [5], [12], possibly resulting in less exploration. Researchers [2], [5] noticed the exploration issue and tried some exploration mechanisms such as $\epsilon$-*greedy* and *softmax*, but neither worked for 2048 according to their reports. Thus, it was simply assumed that the stochastic environment provided enough randomness, and left the efficient exploration for 2048 as an open question.

In this paper, we propose to use *optimistic initialization* (OI) to improve the TD methods for 2048. The approach is to optimistically initialize feature weights to large values in order to encourage exploration [9], [13]. Namely, those feature weights rarely adjusted or visited tend to be high, and therefore the value adjustments are often negative, i.e., these weights tend to be reduced. Thus, agents tend to explore the less visited states next time. All feature weights eventually converge after sufficient visits.

Our experiments show that both TD and TC learning with OI significantly improve the performance. With additional tunings such as expectimax search, multistage learning, and tile-downgrading technique, our design outperforms the previous SOTA results [5] and achieves new SOTA performance, namely an average score of 625 377 and a rate of 72% reaching 32768-tiles. Even more, our method requires only 20% of network weights compared with the previous SOTA method. In addition, for sufficiently large tests, 65536-tiles are reached at

Manuscript received October 9, 2020; revised April 23, 2021; revised July 5, 2021. Date of current version July 5, 2021. This research was supported in part by the Ministry of Science and Technology (MOST) of the Republic of China (Taiwan) under Grant 108-2634-F-009-011, 109-2634-F-009-019, and 110-2634-F-009-022 through Pervasive Artificial Intelligence Research (PAIR) Labs, and the computing resource was supported in part by National Center for High-performance Computing (NCHC) of Taiwan. *(Corresponding author: I-Chen Wu.)*

Hung Guei is with the Department of Computer Science, National Yang Ming Chiao Tung University, Hsinchu, Taiwan (e-mail: hguei@cs.nctu.edu.tw).

Lung-Pin Chen is with the Department of Computer Science, Tunghai University, Taichung, Taiwan (e-mail: lbchen@thu.edu.tw).

I-Chen Wu is with the Research Center for IT Innovation, Academia Sinica, Taipei, and the Department of Computer Science, National Yang Ming Chiao Tung University, Hsinchu, Taiwan (e-mail: icwu@csie.nctu.edu.tw).

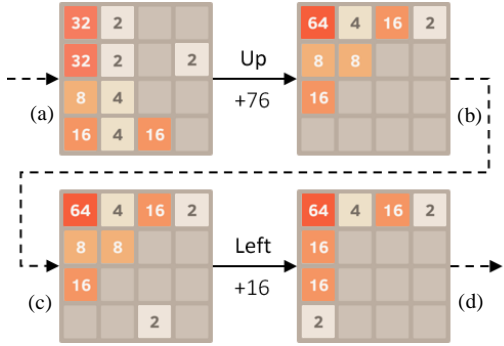

Fig. 1. A segment of a gameplay episode of 2048. The player first slides the puzzle up in (a), merging some tiles and receiving 76 points to (b). After the environment generates a new tile from (b) to (c), the player continues to play by sliding the puzzle left from (c) to (d) with receiving 16 points.

a rate of 0.02%.

The paper is organized as follows. Section II reviews the rules and the related techniques for 2048. Section III introduces the optimistic initialization and the optimistic methods for 2048. Section IV conducts experiments and analyses of the optimistic methods. Section V discusses potentially related techniques with directions of possible future research. Section VI summarizes the results and makes concluding remarks.

## II. BACKGROUND

In this section, the game of 2048 as well as its related methods and techniques are introduced. Section II-A introduces the rules. Section II-B, II-C, and II-D review the foundations and implementations of TD methods. Section II-E reviews the $n$-tuple networks. Section II-F reviews the tree search.

### A. Rules of 2048

*2048* is a single-player stochastic slide-and-merge puzzle game with the objective of sliding the puzzle to merge small tiles into large tiles to create a *2048-tile*. The game is played on a puzzle with 4×4 cells, starting with two randomly placed tiles. Cells on the puzzle are either empty cells or tiles numbered with powers of 2, such as 2-tiles, 4-tiles, 65536-tiles.

Whenever the player slides the puzzle by choosing a direction from up, down, left, and right, all tiles will be moved in the chosen direction as far as possible, i.e., until they reach either the border or another tile [2]. Illustrations are shown in Fig. 1, such as sliding up from (a) to (b), and sliding left from (c) to (d). Upon sliding the puzzle, two adjacent tiles with the same value in the chosen direction, say both $v$-tiles, will be merged into a single $2v$-tile. The player receives $2v$ points as the reward of merging.

After the player slides the puzzle and merges the tiles, the environment randomly adds a new tile on an empty cell as in Fig. 1 (c) from (b). The newly added tile is either a 2-tile or a 4-tile, with probabilities of 0.9 or 0.1 respectively [2]. Then the player continues to slide the puzzle, repeating the above process until there is no possible direction to move. The player wins the game if a 2048-tile is created, but the game can continue until there is no available sliding direction. The final score of a game is the total cumulative rewards of merging tiles.

### B. Reinforcement Learning

*Reinforcement learning* (RL) is a machine learning method that trains an *agent* how to respond to an *environment* with an objective of maximizing the total outcome [9]. The agent continuously interacts with the environment by performing the *actions* to the current *state*, and the environment responds by providing the corresponding *rewards* and the new states.

*Markov decision process* (MDP) is a mathematical framework for decision-making problems, which is commonly used in reinforcement learning [9]. An MDP is constructed by $\langle \mathcal{S}, \mathcal{A}, \mathcal{P}, \mathcal{R} \rangle$, where $\mathcal{S}$ is the finite set of states; $\mathcal{A}$ is the finite set of actions of the state; $\mathcal{P}: \mathcal{S} \times \mathcal{A} \to \mathcal{S}$ is the state transition function; and $\mathcal{R}: \mathcal{S} \times \mathcal{A} \to \mathbb{R}$ is the immediate reward function. The MDP models problems of how an agent interacts with the environment through a sequence of actions with respect to states and rewards. An episode is a sequence of states and actions starting from the beginning till the end.

The game of 2048 can be well modeled as an MDP, in which the player is considered as an agent who takes actions to the states and receives rewards from the environment. For example, the puzzles illustrated in Fig. 1 (a), (b), (c), (d) can be expressed as $s_t$, $s'_t$, $s_{t+1}$, $s'_{t+1}$ respectively, in an episode from $s_0$ to $s_T$, as follows.

$$s_0 \cdots \to s_t \xrightarrow[r_t]{a_t} s'_t \dashrightarrow s_{t+1} \xrightarrow[r_{t+1}]{a_{t+1}} s'_{t+1} \dashrightarrow \cdots s_T \quad (1)$$

The episode starts with an *initial state* $s_0 \in \mathcal{S}$. At steps $t$, the agent performs actions $a_t \in \mathcal{A}(s_t)$ on states $s_t \in \mathcal{S}$ to transform $s_t$ into *afterstates* $s'_t = \psi(s_t, a_t)$, where $\psi$ is a transition function from states to afterstates, i.e., slides the puzzle. The environment responds rewards $r_t = \mathcal{R}(s_t, a_t)$, and then changes $s'_t$ to next states $s_{t+1} \sim \mathcal{P}(s'_t, s_{t+1})$, i.e., adds a new tile. This process repeats until a *terminal state* $s_T \in \mathcal{S}$ that $\mathcal{A}(s_T) = \emptyset$ is reached, where $T$ refers to the end of the episode.

The objective of the problem of MDPs is to find a policy $\pi$ that decides which action to take for any given state, and maximizes the cumulative rewards [9]. The *state value function* is defined as $V(s_t) = \mathbb{E}[r_t + r_{t+1} + \cdots]$. Therefore, the policy $\pi: \mathcal{S} \to \mathcal{A}$ can be derived from the function as

$$\pi(s_t) = \mathrm{argmax}_{a_t}\left(r_t + \sum_{\forall s_{t+1}} \mathcal{P}(s'_t, s_{t+1}) V(s_{t+1})\right), \quad (2)$$

where $a_t \in \mathcal{A}(s_t)$ and $s'_t = \psi(s_t, a_t)$

### C. Temporal Difference Learning

*Temporal difference* (TD) *learning* is a kind of reinforcement learning method that adjusts the state estimations based in part on other learned estimations [9]. This method has been widely applied to many game-playing programs [14]–[21], and was first applied to 2048 by Szubert and Jaśkowski [2], resulting in the first-ever TD-based program that can reach 2048-tiles.

*TD(0)* is the simplest form of TD learning that adjusts the estimation with only one subsequent reward and estimation [9]. After the agent performs an action and receives a reward $r_t$, the environment will provide the next state $s_{t+1}$ whose value is $V(s_{t+1})$. Thus, $r_t + V(s_{t+1})$, known as the *TD target*, is an estimation of $V(s_t)$. Therefore, the estimation error for $s_t$, called the *TD error*, is calculated as

$$\delta_t = r_t + V(s_{t+1}) - V(s_t). \quad (3)$$

Note that in [9] $V(s_{t+1})$ is weighted by a discount factor that is disregarded for simplicity. Then, the estimation of $s_t$, $V(s_t)$, is adjusted with the TD error and the learning rate parameter $\alpha$ as

$$V(s_t) \leftarrow V(s_t) + \alpha \delta_t. \quad (4)$$

TD(0) adjusts the estimation at the current step based on the current reward and the next estimation. Its general form, *n-step TD*, adjusts the estimation based on $n$ subsequent rewards and a more distant estimation [9]. Furthermore, $n$-step TD can be generalized to *TD($\lambda$)*, which adjusts the current estimation based on all subsequent estimations [9], [22]. However, $n$-step TD and TD($\lambda$) are not used in this paper for simplicity.

The above learning framework is to evaluate the state values, i.e., $V(s_t)$. However, from the perspective of taking actions, it is more efficient for 2048 to evaluate the afterstate values, i.e., $V(s'_t)$, instead, called the *afterstate learning framework* [2], [5]. With afterstate values, the policy function can be more efficient than that in (2), as described as follows.

$$\pi(s_t) = \text{argmax}_{a_t}(r_t + V(s'_t)). \quad (5)$$

Similarly, the TD error is then calculated as

$$\delta_t = r_{t+1} + V(s'_{t+1}) - V(s'_t). \quad (6)$$

Depending on the order of adjusting afterstate values within an episode, *forward update* and *backward update* are both common implementations [23], [24], in which the performance of using the latter can be slightly better. In addition, *Q-learning* can also be employed to 2048. However, for 2048 programs with $n$-tuple networks, a Q-learning implementation is complicated, and its performance is significantly worse than TD learning [2], so it has not been widely used.

*D. Advanced TD Methods*

*Multistage temporal difference* (MS-TD) *learning* proposed by Yeh *et al.* [3] is a kind of hierarchical TD learning [25] that divides the entire episode into multiple stages, in which each stage has an independent value function, used in [3], [5], [10], [26]. MS-TD learning improves the performance at the cost of additional storage for stages. The work [3] applied this method with 3-stage to 2048 and obtained the first-ever 65536-tile.

*Temporal coherence* (TC) *learning* proposed for 2048 by Jaśkowski [5] is a TD variant with adaptive learning rates [12]. Instead of adjusting the learning rate $\alpha$ directly, this method introduces new parameters $\beta_i$ for the $i^{\text{th}}$ feature weight, denoted by $\theta_i$, to modulate the adjustments as

$$\theta_i \leftarrow \theta_i + \alpha \beta_i \delta_t. \quad (7)$$

$\beta_i$ represents the *coherence* of $\theta_i$, and is calculated from two parameters $E_i$ and $A_i$ for each weight, as

$$\beta_i = \begin{cases} |E_i|/A_i, & \text{if } A_i \neq 0 \\ 1, & \text{otherwise.} \end{cases} \quad (8)$$

Both $E_i$ and $A_i$ are initialized with 0 and adjusted by

$$E_i \leftarrow E_i + \delta_t \text{ and } A_i \leftarrow A_i + |\delta_t|. \quad (9)$$

Therefore, the amount of adjustment is automatically reduced by coherence $\beta_i$. TC learning is an effective learning rate decay method with an overhead of triple required memory. This technique can be integrated with MS-TD, e.g., a 16-stage TC method was able to achieve 609 104 points and a 70% chance of reaching 32768-tiles on average [5].

*E. N-tuple Networks*

A straightforward method to estimate state values $V(s)$ is to use a tabular implementation for the whole state space. However, the state space requires $18^{(4\times4)}$ for 2048, too large to be implemented. Therefore, a *function approximator* is applied in practice. *N-tuple network* is a function approximator that has been successfully applied to applications such as Connect4 [20], Othello [21], pattern recognition [27], as well as 2048 [2], [3], [5], [8].

For $n$-tuple networks for 2048, an *n-tuple* $\phi$ is a sequence of distinct features, each representing a tuple of $n$ designated cells on the puzzle. For example, let $\phi_{R1}$ be a 4-tuple that denotes features from the first row. For a puzzle $s$, say the one in Fig. 1 (a), a *feature* $\phi_{R1}(s)$ refers to (32, 2, 0, 0). Similarly, for $s'$, say the one in Fig. 1 (b), a feature $\phi_{R1}(s')$ refers to (64, 4, 16, 2).

An $n$-tuple network is an implementation for a set of weights of $n$-tuple features. In order to access these *feature weights*, let $\phi$ be an $n$-tuple associated with a lookup table LUT in which the feature weight of $\phi(s)$ is stored at a distinct LUT[$\phi(s)$]. An illustration of the implementation for $\phi_{R1}$ is as follows. The lookup table LUT$_{R1}$ for $\phi_{R1}$ needs to contain $c^4$ distinct feature weights, since $\phi_{R1}$ consists of 4 cells, each with $c$ distinct cell values. Intrinsically, $c$ is 18, i.e., from empty cell to 131072-tile, but is usually set to 16 or 17 for efficiency, since 65536-tile and 131072-tile are rarely obtained.

In this paper, we define an *$m\times n$-tuple network* to be $m$ different $n$-tuples $\phi_1, \ldots, \phi_m$ with their corresponding lookup tables LUT$_1, \ldots,$ LUT$_m$. Given a state $s$, the state value estimation $V(s)$, is calculated by summing all of the $m$ feature weights LUT$_i[\phi_i(s)]$ of state $s$ as

$$V(s) = \sum_{i=1}^{m} \text{LUT}_i[\phi_i(s)]. \quad (10)$$

When adjusting a state estimation $V(s)$ by a TD error $\delta$, the adjustment is equally distributed to $m$ feature weights of state $s$. Equation (11) shows how $\delta$ is distributed to a feature weight LUT$_i[\phi_i(s)]$ in terms of TD(0) with learning rate $\alpha$:

$$\text{LUT}_i[\phi_i(s)] \leftarrow \text{LUT}_i[\phi_i(s)] + (\alpha/m)\delta. \quad (11)$$

Note that only the $m$ feature weights corresponding to state $s$ need to be adjusted, with the same adjustment $(\alpha/m)\delta$.

For 2048, *symmetric sampling* is a widely used technique that shares feature weights of tuples eight times by rotating and mirroring [2], [3], [5], [8]. A *symmetrically sampled $m\times n$-tuple network* involves $8m$ $n$-tuples actually, which improves the overall performance without additional lookup tables. For example, let $\phi_{R1'}$ be a 4-tuple produced by rotating $\phi_{R1}$ counterclockwise, then, its feature $\phi_{R1'}(s)$ is (16, 8, 32, 32). Both $\phi_{R1}$ and $\phi_{R1'}$ share the same lookup table LUT$_{R1}$. Note that an $m\times n$-tuple network refers to a symmetrically sampled $m\times n$-tuple network in the rest of this paper for simplicity.

Designing an effective $m\times n$-tuple network is not trivial, and has been investigated as a research topic for 2048 [2], [3], [8], [11]. Fig. 2 (a), (b), (d), and (e) illustrate the 4×6-tuple network proposed by Yeh *et al.* [3], [10]; all 5 6-tuples illustrate the 5×6-tuple network used by Jaśkowski [5]. Fig. 3 illustrates a full set of 6-tuples proposed by Matsuzaki [8], where the best $k$×6-tuple network consists of the first $k$ listed 6-tuples, e.g., the 8×6-tuple network contains all from Fig. 3 (a) to (h).

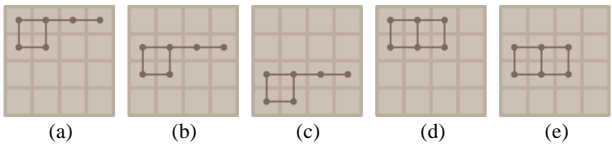

Fig. 2. The 4×6-tuple network proposed by Yeh *et al.* [3]: (a), (b), (d), and (e); and the 5×6-tuple network used by Jaśkowski [5]: (a)–(e).

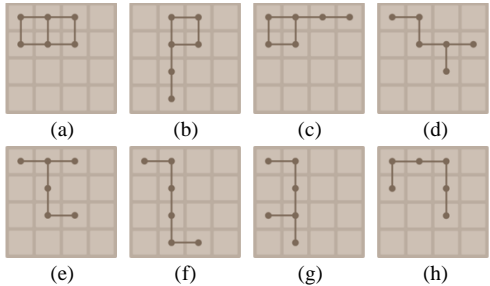

Fig. 3. The 8×6-tuple network proposed by Matsuzaki [8]: (a)–(h).

### F. Expectimax Search

*Expectimax search* is a technique for stochastic games whose game tree is composed of *max nodes* and *chance nodes* [28]–[30], corresponding to states and afterstates respectively. Namely, a max node of state $s_t$ searches all its afterstates $s'_t$ to find the best action; and a chance node of afterstate $s'_t$ is evaluated based on either the expected value of next states, or the TD value $V(s'_t)$ when the ply limit is reached. For 2048 with afterstate learning framework, a $p$-ply fixed-depth search tree has at most $p$ layers of chance nodes.

To avoid redundant search, the search is usually integrated with a *transposition table*, which caches the previously seen states and associated values. A transposition table for 2048 is implemented using hashing techniques such as *Zobrist hashing* [31], [32] and *MurmurHash* [33], [34].

## III. OPTIMISTIC METHODS

In this section, optimistic methods for 2048 are introduced. Section III-A reviews the potential problems of insufficient exploration in previous works. Sections III-B and III-C present TD learning with optimistic initialization for 2048. Finally, Section III-D describes how to determine initial values and use them in $n$-tuple networks.

### A. Insufficient Exploration

In reinforcement learning, the *exploration-exploitation dilemma* and has been intensively studied by researchers for decades [9], [13], [35]. Many methods, including UCB, $\epsilon$-greedy, softmax, as well as optimistic initialization, are proposed to balance between exploration and exploitation [36] for better learning performance.

Most previous RL related works for 2048 [2], [3], [5], [6], [8], [10], [11], [23], [26] are based on simple TD method or some of its variant. Among these works, the agents simply follow the greedy policy with respect to the estimations, i.e., always select an action with maximum value during training. Past works [2], [5] have already noticed this issue, and have tried some explicit exploration techniques such as $\epsilon$-greedy and softmax. However, they did not succeed to have these techniques work for 2048. As a result, these TD methods involve no explicit exploration to choose actions and fully rely on the stochasticity of the environment for exploration.

For example, MS-TD learning is proposed to cope with the issue that TD learning tends not to reach large tiles [3], [10]. MS-TD learning improves performance by employing additional networks, while the issue of potential exploration deficiency of each stage remains not addressed. Another effective technique is TC learning, which reduces weight adjustments to automatically accelerate network convergence [5]. This approach is not for improving exploration, but for improving exploitation with fast convergence. However, fast convergence exacerbates the exploration issue, especially when applying a large $n$-tuple network.

### B. Optimistic Initialization

*Optimistic initialization* (OI) is an approach that employs optimistic initial estimations to encourage exploration [9], [13]. This approach has been widely applied to RL applications, and is considered to have good convergence in practice [9]. Instead of setting the estimations to zero or random, the technique optimistically initializes them to a large value to encourage the agent to explore. Due to the large value, the estimations tend to be reduced once the corresponding states are visited, therefore leading an agent to select unexplored or rarely explored actions the next time when it revisits the same state. This process repeats until all the actions are sufficiently explored, even if the greedy policy is always applied during the training.

The estimations will eventually converge and may even converge to a near-optimal policy when the initial value is set sufficiently large [35], [37], namely, the upper bound of the value function. Thus, the learning algorithm needs to explore unvisited or rarely visited states and reduce the corresponding estimations before exploiting the best one [38]. Therefore, OI may significantly increase the training time in exchange for encouraging exploration.

### C. Optimistic TD Methods

Past research summarized that non-greedy behaviors, e.g., $\epsilon$-greedy, significantly inhibit the learning performance [2], [5]. In this paper, we propose to use OI as an exploration mechanism for 2048. The proposed OI methods perform exploration while conserving the greedy behavior, therefore mitigating the previous inhibition phenomenon. In addition, since the explicit exploration technique is independent of some existing learning methods such as multistage and TC learning, OI can be easily applied together with these methods.

In this paper, the first objective is to improve the existing TD and TC methods for 2048 by using OI, which forms the *optimistic TD* (OTD) and the *optimistic TC* (OTC) *learning*, respectively. Second, we propose a two-phase optimistic method, called *OTD+TC learning*, a hybrid learning paradigm that combines the advantages of both methods. OTD+TC learning first performs TD learning with a fixed learning rate to further encourage exploration for a while, and then, in the second phase, continues with TC fine-tuning for exploitation.

TABLE I
PERFORMANCE OF OTD METHODS IN THE 4×6-TUPLE NETWORK

| $V_{init}$ | Average Score | 8192 [%] | 16384 [%] | 32768 [%] |
|---|---|---|---|---|
| 0 | 251 920 ± 11 217 | 92.87 ± 0.90% | 65.23 ± 2.86% | 0.00 ± 0.00% |
| 40k | 254 784 ± 13 034 | 93.20 ± 1.08% | 65.89 ± 3.60% | 0.00 ± 0.00% |
| 80k | 253 393 ± 11 732 | 92.52 ± 1.13% | 65.25 ± 3.35% | 0.00 ± 0.00% |
| 160k | 244 875 ± 10 590 | 91.40 ± 1.41% | 62.29 ± 3.27% | 0.00 ± 0.00% |
| 320k | 239 075 ± 14 781 | 89.68 ± 2.11% | 59.73 ± 5.00% | 0.00 ± 0.00% |
| 640k | 228 032 ± 1056 | 87.64 ± 0.44% | 55.63 ± 0.48% | 0.00 ± 0.00% |

TABLE II
PERFORMANCE OF OTC METHODS IN THE 4×6-TUPLE NETWORK

| $V_{init}$ | Average Score | 8192 [%] | 16384 [%] | 32768 [%] |
|---|---|---|---|---|
| 0 | 265 481 ± 3318 | 93.36 ± 1.07% | **68.33 ± 0.52%** | 0.00 ± 0.00% |
| 40k | 263 284 ± 2445 | **93.68 ± 0.32%** | 66.72 ± 1.79% | 0.00 ± 0.00% |
| 80k | 265 703 ± 2471 | 93.60 ± 0.26% | 67.93 ± 1.49% | 0.00 ± 0.00% |
| 160k | 264 058 ± 1412 | 93.24 ± 0.28% | 67.47 ± 0.35% | 0.00 ± 0.00% |
| 320k | **280 281 ± 5267** | 93.11 ± 0.53% | 67.83 ± 1.50% | **5.30 ± 0.58%** |
| 640k | 252 254 ± 6836 | 89.18 ± 1.48% | 60.22 ± 1.98% | 3.25 ± 0.38% |

When compared to TD or TC, OTD+TC learning includes two new hyperparameters, $V_{init}$ and $P_{TC}$: $V_{init}$ is the initial value of the function approximator; and $P_{TC}$ is the proportion of TC fine-tuning phase to the total phases. If $V_{init}$ is set to 0, OTD+TC becomes non-optimistic. If $P_{TC}$ is set to 0% (or 100%), OTD+TC becomes pure OTD (or OTC).

### D. Initial Values for OI

Based on the proof of the optimistic Q-learning in [37], the initial value $V_{init}$ should be set to the theoretical maximum to ensure that the network converges to a near-optimal policy. However, the theoretical maximum for 2048 is an extremely large number that is nearly impossible to obtain. Using such a large $V_{init}$ wastes too much time, it is non-trivial to choose $V_{init}$ such that training result and training time are balanced.

In this paper, the initial value $V_{init}$ is chosen as illustrated as follows. Consider using an $m \times n$-tuple network for training with non-optimistic TD learning, and set $V_{init}$ an estimated average score. To initialize the network feature weights, we evenly distribute the value $V_{init}$ over these feature weights as

$$\text{LUT}[i] \leftarrow V_{init}/m \text{ for all } i. \quad (12)$$

## IV. EXPERIMENTS

In this section, experiments are presented to analyze the effectiveness of optimistic TD learning for 2048. Common training settings are described as follows. For convergence, each $n$-tuple network was trained with 100M episodes by using the afterstate learning framework. The learning performance was evaluated every 1M training episodes, and each performance evaluation consists of 100k testing episodes. For statistics, each method was trained with five runs, each with one individual trained network with a different initial random seed.

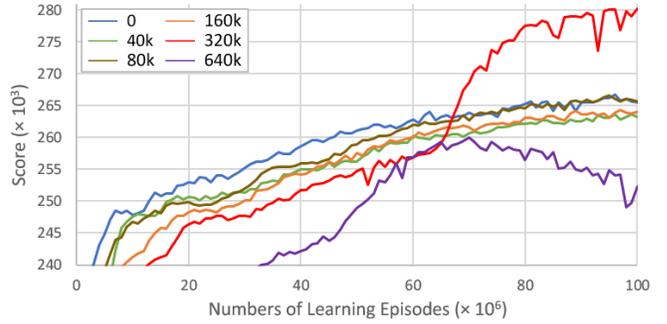

Fig. 4. Average scores of OTC methods in the 4×6-tuple network.

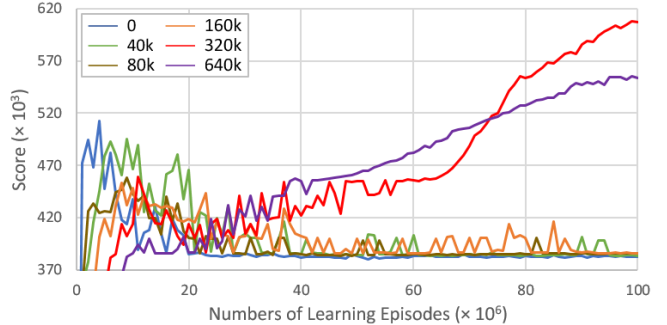

Fig. 5. Maximum scores of OTC methods in the 4×6-tuple network.

The experiments were performed on workstations with Intel Xeon E5 processors. To speed up the training process, 20 threads *lock-free optimistic parallelism* was applied as in [5]. Specific settings will be described in each experiment below.

The results are presented in tables and figures in which each value represents the average of five trained networks. For all tables (e.g., Table I), a value indicates the average result after 100M training episodes and is accomplished with the 95% confidence interval. For all figures (e.g., Fig. 4), a point at time $t$ indicates the average result after $t$M training episodes.

The experiments are organized as follows. Section IV-A analyzes the effectiveness of initial values on OTD and OTC learning. Section IV-B analyzes OTD+TC learning with different fine-tuning proportions on networks of different sizes. Section IV-C further improves the performance by fine-tuning with other techniques for OTD+TC learning.

### A. OTD and OTC Learning

In order to demonstrate the effectiveness of OI and obtain an appropriate initial value, we analyze how OTD and OTC learning perform with various initial values. Two different $n$-tuple networks, Yeh's 4×6-tuple (Fig. 2) and Matsuzaki's 8×6-tuple (Fig. 3), were used. Based on their average scores with non-optimistic TD learning, we test the initial value $V_{init}$ to 0, 40k, 80k, 160k, 320k, 640k, respectively. For OTD learning, the learning rate $\alpha$ was initially set to 0.1 and was reduced to 0.01 and 0.001 after completing 50% and 75% of training; for OTC learning, a larger initial learning rate is required, so $\alpha$ was set to 1.0 without manual decay as in [5]. The results are organized as follows. OTD and OTC learning for the 4×6-tuple will be discussed first, and then followed by OTD and OTC learning for the 8×6-tuple.

TABLE III
PERFORMANCE OF OTD METHODS IN THE 8×6-TUPLE NETWORK

| $V_{init}$ | Average Score | 8192 [%] | 16384 [%] | 32768 [%] |
|---|---|---|---|---|
| 0 | 309 208 ± 5788 | **97.24 ± 0.58%** | 85.13 ± 0.78% | 0.00 ± 0.00% |
| 40k | 306 438 ± 2400 | 97.27 ± 0.11% | 84.11 ± 0.44% | 0.00 ± 0.00% |
| 80k | 365 364 ± 4856 | 97.16 ± 0.17% | 85.59 ± 0.67% | 21.23 ± 1.04% |
| 160k | **369 172 ± 3341** | 97.18 ± 0.30% | **85.70 ± 0.63%** | 22.47 ± 1.31% |
| 320k | 361 471 ± 5473 | 96.77 ± 0.28% | 84.41 ± 0.96% | 21.75 ± 0.74% |
| 640k | 339 492 ± 6747 | 95.79 ± 0.28% | 81.80 ± 1.07% | 18.11 ± 2.12% |

TABLE IV
PERFORMANCE OF OTC METHODS IN THE 8×6-TUPLE NETWORK

| $V_{init}$ | Average Score | 8192 [%] | 16384 [%] | 32768 [%] |
|---|---|---|---|---|
| 0 | 310 259 ± 4056 | 96.54 ± 0.13% | **84.81 ± 0.43%** | 0.00 ± 0.00% |
| 40k | 310 347 ± 6201 | 96.72 ± 0.71% | 83.59 ± 1.38% | 0.00 ± 0.00% |
| 80k | 311 214 ± 2205 | **97.07 ± 0.22%** | 83.45 ± 0.23% | 0.00 ± 0.00% |
| 160k | **361 298 ± 4833** | 96.34 ± 0.20% | 84.12 ± 0.52% | 22.26 ± 2.12% |
| 320k | 360 228 ± 18 829 | 95.73 ± 0.39% | 82.28 ± 2.25% | **22.74 ± 2.38%** |
| 640k | 325 357 ± 33 416 | 92.76 ± 2.55% | 75.88 ± 6.10% | 17.71 ± 4.89% |

TABLE V
DIFFERENCE BETWEEN OTD, OTC, AND OTD+TC METHODS IN N-TUPLE NETWORKS OF DIFFERENT SIZES

| Network | OTD | OTD+TC | OTC |
|---|---|---|---|
| 4×6-tuple | 239 075 ± 14 781 (0.00 ± 0.00%) | 261 433 ± 3143 (0.02 ± 0.02%) | **280 281 ± 5267 (5.30 ± 0.58%)** |
| 5×6-tuple | 272 904 ± 7788 (0.01 ± 0.01%) | 279 653 ± 8707 (0.47 ± 1.08%) | **313 948 ± 2433 (11.48 ± 0.97%)** |
| 6×6-tuple | 314 324 ± 8883 (8.33 ± 5.58%) | **337 951 ± 4655 (14.39 ± 0.96%)** | 324 714 ± 17 688 (12.88 ± 5.88%) |
| 7×6-tuple | 352 986 ± 3553 (18.43 ± 0.69%) | **360 286 ± 2332 (19.18 ± 0.70%)** | 335 042 ± 23 023 (16.56 ± 6.61%) |
| 8×6-tuple | 361 471 ± 5473 (21.75 ± 0.74%) | **370 907 ± 1630 (22.18 ± 0.53%)** | 360 228 ± 18 829 **(22.74 ± 2.38%)** |

Networks initialized with $V_{init}$ = 320k; OTD+TC used $P_{TC}$ = 10%.

For the 4×6-tuple network, OTD learning did not obviously improve the learning quality, as shown in Table I. On the other hand, OTC learning significantly improved the performance, especially for the reaching rate of 32768-tiles. As the results summarized in Table II and Fig. 4, OTC with 320k outperformed non-optimistic TC ($V_{init}$ = 0), and even achieved a rate of 5.30% reaching 32768-tiles, which is the first-ever reaching rate with such a 1-stage network. As shown in Fig. 5, many methods reached maximum scores around 385 000. This implicitly indicates to achieve states with 16384-tile, 8192-tile, 4096-tile, and 2048-tile. Note that a method can continue to obtain much more scores if a 32768-tile is reached. Only OTC with 320k and 640k provided enough exploration to achieve 32760-tiles. Interestingly, as shown in Fig. 4, OTC with 640k reached 32768-tiles earlier than OTC with 320k, but eventually ended at a worse average score. In general, a large $V_{init}$ needs more time to converge. Our conjecture is that 640k is too large to converge. Thus, we prefer 320k for obtaining higher average scores.

For the 8×6-tuple network, in contrast to the 4×6-tuple, OTD learning significantly improves the performance. As shown in Table III and Fig. 6, networks initialized with appropriate values (80k ≤ $V_{init}$ < 640k) achieved higher scores, and even reached 32768-tiles well, especially for $V_{init}$ = 160k and 320k. As shown in Fig. 7, networks trained with low exploration ($V_{init}$ < 80k) still stuck at the barrier of 32768-tiles; and networks trained with 640k still did not converge. On the other hand, the results of OTC learning are summarized in Table IV. OTC learning also performed well, but its performance was slightly worse than OTD for the 8×6-tuple network. Interestingly, for $V_{init}$ = 80k on the 8×6-tuple network, OTD achieved 32768-tiles, while OTC did not. Since such a phenomenon did not occur on the 4×6-tuple, it can be derived from this observation that exploration may be restricted by TC method since the weight adjustments drop too fast, therefore, a pure TD method plays an important role for larger networks.

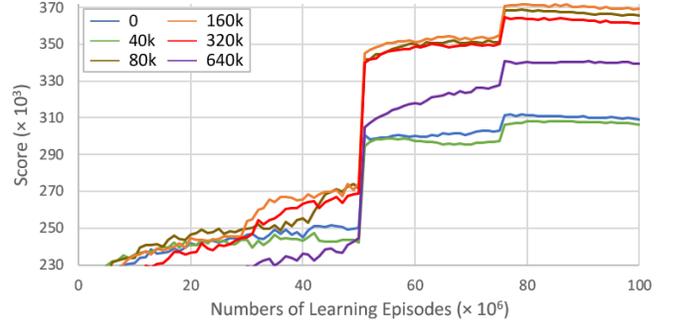

Fig. 6. Average scores of OTD methods in the 8×6-tuple network.

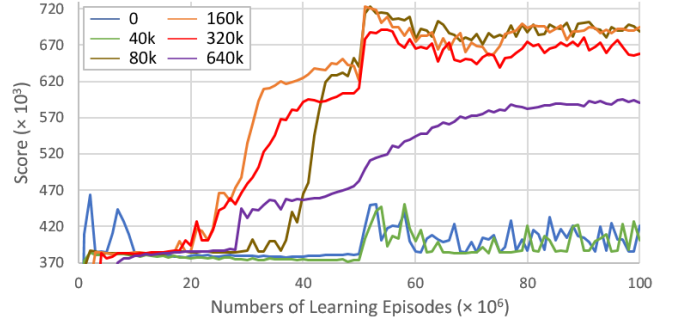

Fig. 7. Maximum scores of OTD methods in the 8×6-tuple network.

*B. OTD+TC Learning*

This experiment analyzes OTD+TC learning. In order to analyze the correlation between OI and network size in detail, five experimental network sizes were chosen, and were divided into two classes: the smaller networks, including Yeh's 4×6-tuple and 5×6-tuple (Fig. 2); and the larger networks, including Matsuzaki's 6×6-tuple, 7×6-tuple, and 8×6-tuple (Fig. 3). Since 320k performed well in most cases as described above, $V_{init}$ is set to 320k in the rest of the experiments.

First, we would like to analyze the effectiveness of OTD+TC learning. The average scores and the 32768-tile reaching rates of OTD, OTC, and OTD+TC with $P_{TC}$ = 10% in the five networks are summarized in Table V. OTD+TC outperforms OTD in all cases, and slightly outperforms OTC in most cases of using larger networks. In the cases of using smaller networks, OTC outperforms OTD+TC; while OTD+TC hardly obtained 32768-tiles. In brief, OTC learning performed well regardless of the network size, and OTD+TC learning outperforms OTC learning in most larger networks.

Second, it is interesting to investigate how the fine-tuning proportions $P_{TC}$ affects the performance. Table VI summarizes the results of more $P_{TC}$ of 10%, 20%, 30%, 50%, 70%, and 90%

TABLE VI
PERFORMANCE OF OTD+TC METHODS IN THE 8×6-TUPLE NETWORK

| $P_{TC}$ | Average Score | 8192 [%] | 16384 [%] | 32768 [%] |
|---|---|---|---|---|
| 10% | 370 907 ± 1630 | **97.26 ± 0.16%** | 85.43 ± 0.34% | 22.18 ± 0.53% |
| 20% | **371 198 ± 8301** | 97.19 ± 0.53% | **85.55 ± 1.21%** | **22.78 ± 1.39%** |
| 30% | 366 868 ± 7767 | 97.07 ± 0.45% | 84.82 ± 1.28% | 22.28 ± 1.38% |
| 50% | 355 806 ± 13 994 | 96.32 ± 0.73% | 83.64 ± 1.97% | 20.61 ± 2.24% |
| 70% | 344 570 ± 20 316 | 95.99 ± 1.06% | 82.04 ± 3.58% | 17.78 ± 2.83% |
| 90% | 298 673 ± 18 995 | 95.25 ± 1.64% | 80.05 ± 4.60% | 0.86 ± 2.00% |

Networks were trained with initial value $V_{init}$ = 320k.

TABLE VII
PERFORMANCE OF OPTIMISTIC TD METHODS TOGETHER WITH EXPECTIMAX SEARCH IN THE 4×6-TUPLE AND THE 8×6-TUPLE NETWORKS

| Network | Search | Average Score | 8192 [%] | 16384 [%] | 32768 [%] |
|---|---|---|---|---|---|
| 4×6-tuple | 1-ply | 280 123 ± 4920 | 93.00 ± 0.61% | 67.79 ± 1.39% | 5.32 ± 0.43% |
| | 3-ply | 417 712 ± 2988 | 99.66 ± 0.10% | 94.95 ± 0.60% | 37.18 ± 1.62% |
| | 5-ply | 445 085 ± 8843 | 99.80 ± 0.89% | 96.60 ± 5.02% | 51.20 ± 3.58% |
| 8×6-tuple | 1-ply | 370 194 ± 4366 | 97.23 ± 0.18% | 85.30 ± 0.46% | 22.18 ± 0.92% |
| | 3-ply | 475 126 ± 6407 | 99.73 ± 0.08% | 96.88 ± 0.78% | 50.73 ± 1.56% |
| | 5-ply | **500 098 ± 8590** | **100.00 ± 0.00%** | **98.20 ± 3.85%** | **57.80 ± 2.97%** |

The 4×6-tuple and the 8×6-tuple were trained with OTC and OTD+TC, respectively.

TABLE VIII
PERFORMANCE OF MULTISTAGE OPTIMISTIC TD METHODS TOGETHER WITH EXPECTIMAX SEARCH IN THE 4×6-TUPLE AND THE 8×6-TUPLE NETWORKS

| Network | Search | Average Score | 8192 [%] | 16384 [%] | 32768 [%] |
|---|---|---|---|---|---|
| 2-stage 4×6-tuple ($V_{init}$=0) | 1-ply | 291 428 ± 5229 | 93.30 ± 1.21% | 68.58 ± 1.05% | 7.72 ± 1.80% |
| | 3-ply | 463 784 ± 12 569 | 99.72 ± 0.15% | 96.16 ± 0.36% | 46.97 ± 1.28% |
| | 5-ply | 478 831 ± 13 174 | **100.00 ± 0.00%** | 98.40 ± 1.10% | 53.80 ± 3.29% |
| 2-stage 4×6-tuple ($V_{init}$=320k) | 1-ply | 290 549 ± 4105 | 93.03 ± 0.62% | 67.88 ± 1.36% | 8.42 ± 0.44% |
| | 3-ply | 474 167 ± 7087 | 99.70 ± 0.15% | 95.41 ± 0.98% | 46.02 ± 2.53% |
| | 5-ply | 485 588 ± 30 802 | **100.00 ± 0.00%** | 97.00 ± 4.00% | 53.00 ± 9.70% |
| 2-stage 8×6-tuple ($V_{init}$=0) | 1-ply | 386 078 ± 17 104 | 97.37 ± 0.23% | 85.25 ± 1.23% | 26.67 ± 1.64% |
| | 3-ply | 506 486 ± 50 356 | 99.66 ± 0.11% | 96.89 ± 0.32% | 55.50 ± 3.07% |
| | 5-ply | 516 761 ± 67 952 | **100.00 ± 0.00%** | 97.40 ± 3.35% | 56.80 ± 8.53% |
| 2-stage 8×6-tuple ($V_{init}$=320k) | 1-ply | 404 288 ± 2583 | 97.25 ± 0.11% | 85.37 ± 0.39% | 30.17 ± 0.48% |
| | 3-ply | 538 582 ± 5692 | 99.70 ± 0.12% | 96.83 ± 0.32% | 57.58 ± 0.81% |
| | 5-ply | 581 896 ± 14 765 | 99.80 ± 0.89% | **98.60 ± 2.68%** | 66.40 ± 6.42% |
| 3-stage 8×6-tuple ($V_{init}$=320k) | 1-ply | 412 492 ± 3666 | 97.23 ± 0.13% | 85.34 ± 0.35% | 33.55 ± 0.91% |
| | 3-ply | 545 231 ± 6479 | 99.66 ± 0.12% | 96.90 ± 0.44% | 60.48 ± 0.19% |
| | 5-ply | **588 426 ± 30 723** | 99.80 ± 0.89% | 98.20 ± 2.19% | **72.80 ± 8.17%** |

The 4×6-tuple and the 8×6-tuple were trained with OTC and OTD+TC, respectively.

for the 8×6-tuple network. Fig. 8 and Fig. 9 illustrates the learning curves of 10%, 50%, 70%, 90%, with also OTD and OTC as baselines. Based on the results, it is observed that high proportions (e.g., 70% and 90%) limit the ability to explore, and lead to worse results. Therefore, OTD+TC learning should be performed with low fine-tuning proportions such as 10% or 20% to provide sufficient exploration, as well as prevent potential overfitting issue in practice.

### C. Further Improvements

In order to further improve the performance, we demonstrate how to apply optimistic methods together with other established techniques, such as expectimax search, multistage learning, and tile-downgrading.

#### 1) Expectimax Search

As also described in [2], [5], expectimax search can be used to further improve the performance. We applied expectimax search to two cases, the 4×6-tuple network with OTC and the 8×6-tuple with OTD+TC, since both performed generally well as shown above. The performance of three different depths, 1-ply, 3-ply, and 5-ply, were evaluated for the tests of 1M, 10k, and 100 episodes, respectively. Each evaluation is also repeated five times by using the five trained networks. Transposition tables were used to speed up the search.

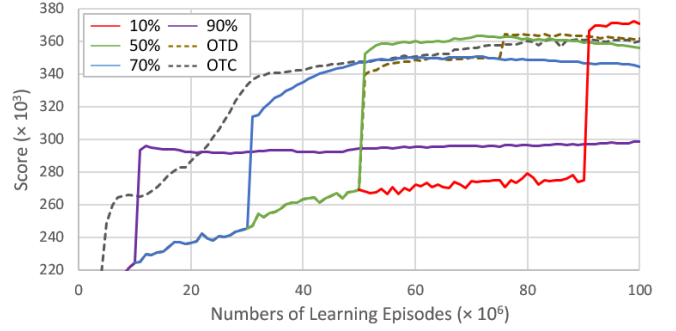

Fig. 8. Average scores of OTD+TC methods in the 8×6-tuple network.

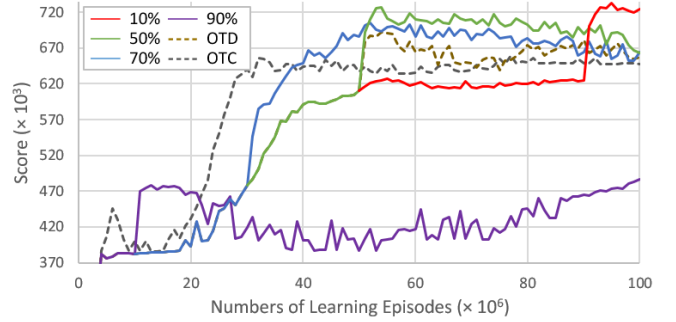

Fig. 9. Maximum scores of OTD+TC methods in the 8×6-tuple network.

Table VII lists the performance of the 4×6-tuple and the 8×6-tuple network with 1-ply, 3-ply, and 5-ply expectimax search. Benefited from OI, both average scores and 32768-tile reaching rates achieved SOTA with respect to the same 1-stage networks and search depths, to the best of our knowledge. However, the performance was saturated around 5-ply.

#### 2) Multistage Learning

We further evaluate optimistic TD learning together with multistage paradigm [3], named *multistage OTD+TC* (MS-OTD+TC) learning. As in Section IV-C1, we also use the two cases, the 4×6-tuple network with OTC and the 8×6-tuple network with OTD+TC. For both, the first stages started with initial states, and the second stages started with states with 16384-tile. Note that the first stages are directly copied from trained networks, and the second stages also received 100M training episodes. All stages were trained using their appropriate optimistic methods and hyperparameters, $V_{init}$ and $P_{TC}$, as mentioned in previous experiment.

Table VIII presents the performance of the 2-stage 4×6-tuple and 8×6-tuple network with 1-ply, 3-ply, and 5-ply expectimax search. For comparison, the non-optimistic 2-stage networks are also provided as baselines. In the second stage, average scores were improved by around 10 ± 5%. For further improvement, we attempted to add the third stage to the 2-stage 8×6-tuple network by starting with 16384-tile + 8192-tile. However, such a 3-stage 8×6-tuple network only improved the average scores by less than 2%, as presented in Table VIII. Our conjecture is that the hyperparameters $V_{init}$ and $P_{TC}$ for the 8×6-tuple, tuned for the first stage, did not fit the third stage.

TABLE X
COMPARISON BETWEEN SOTA OPTIMISTIC METHODS AND THE PREVIOUS SOTA METHOD

| Authors | Methods | Weights | Search | Average Score | 8192 [%] | 16384 [%] | 32768 [%] | # Games |
|---|---|---|---|---|---|---|---|---|
| Jaśkowski [5] | MS-TC 16-stage 5×6-tuple network | 1342.2M | 1-ply | 324 710 ± 11 043 | 90% | 68% | 19% | 1000 |
| | | | 3-ply | 511 759 ± 12 021 | 99% | 92% | 50% | 1000 |
| | | | 5-ply | 545 833 ± 21 500 | **100%** | 97% | 54% | 300 |
| | | | 1000ms | 609 104 ± 38 433 | 98% | 97% | 70% | 100 |
| This work | MS-OTC 2-stage 5×6-tuple network with tile-downgrading | 167.8M | 1-ply | 331 073 ± 6699 | 94.22 ± 0.12% | 74.41 ± 0.44% | 15.38 ± 0.62% | 1 000 000 |
| | | | 3-ply | 526 178 ± 21 752 | 99.69 ± 0.18% | 95.89 ± 0.77% | 51.07 ± 3.87% | 10 000 |
| | | | 5-ply | 574 245 ± 25 727 | 99.80 ± 0.89% | 97.00 ± 2.00% | 60.60 ± 5.02% | 100 |
| | MS-OTD+TC 2-stage 8×6-tuple network with tile-downgrading | 268.4M | 1-ply | 412 785 ± 2208 | 97.24 ± 0.12% | 85.39 ± 0.35% | 30.16 ± 0.38% | 1 000 000 |
| | | | 3-ply | 563 316 ± 5650 | 99.63 ± 0.19% | 96.88 ± 0.12% | 57.90 ± 1.63% | 10 000 |
| | | | 5-ply | 608 679 ± 42 177 | 99.80 ± 0.89% | 97.80 ± 1.67% | 67.40 ± 12.21% | 100 |
| | | | 6-ply | **625 377 ± 40 936** | 99.80 ± 0.89% | **98.80 ± 0.89%** | **72.00 ± 12.00%** | 100 |

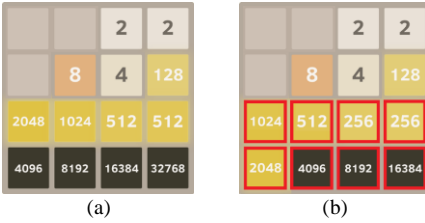

Fig. 10. States before and after tile-downgrading. The original state (a) is translated to its downgraded state (b), where the highlighted tiles indicate that the values have been halved by tile-downgrading.

*3) Tile-Downgrading*

*Tile-downgrading expectimax search* is a technique that searches states by translating them into downgraded states. Let us illustrate it by an example as in Fig. 10, where (a) is an original state and (b) is its downgraded state. In the original state (a), 256-tile is the largest missing tile not exceeding the largest tile, i.e., 32768-tile. This technique is to halve all tile values larger than 256. Thus, the downgraded state is derived as (b). Then, we estimate the value of the downgraded state, instead of the original state. Since states with large tiles have relatively less chance to be trained, their state estimations are relatively less accurate. For the downgraded states, their tile values are relatively smaller, implying higher chances to be trained and higher accuracy.

The tile-downgrading can only be applied to the root state of a search tree when the following two conditions hold in the root: there exists a missing tile like 256-tile as above; and there exists 32768-tiles or larger tiles, which is so chosen since it performed best in most of our experiments. Let a root state $s$ satisfy the two conditions. The tile-downgrading expectimax search is to translate $s$ into a downgraded state $\tilde{s}$ and then use the expectimax search to choose the best action $\tilde{a}$ to play. Note that no more tile-downgrading is required inside the search. Since this technique preserves the puzzle structure while halving some tile values, the action $\tilde{a}$ of $\tilde{s}$ can be directly applied to the original state $s$.

The tile-downgrading expectimax search was evaluated on the 2-stage 8×6-tuple network trained with $V_{init} = 320k$. From Table X, the tile-downgrading method improved the average

TABLE IX
IMPROVEMENT OF INDIVIDUAL TECHNIQUES TO THE BEST RESULT

| Method | Average Score | 32768 [%] |
|---|---|---|
| V0 = 1-stage 8×6-tuple TD ($V_{init}$=0) | 309 208 ± 5788 | 0.00 ± 0.00% |
| V1 = V0 + OI ($V_{init}$=320k) | 361 471 ± 5473 | 21.75 ± 0.74% |
| V2 = V1 + TC fine-tuning ($P_{TC}$=10%) | 370 194 ± 4366 | 22.18 ± 0.92% |
| V3 = V2 + multistage (2-stage) | 404 288 ± 2583 | 30.17 ± 0.48% |
| V4 = V3 + expectimax search (6-ply) | 586 583 ± 17 043 | 65.40 ± 2.28% |
| V5 = V4 + tile-downgrading | **625 377 ± 40 936** | **72.00 ± 12.00%** |
| X1 = V5 − OI | 592 390 ± 39 073 | 63.60 ± 10.35% |
| X2 = V5 − TC fine-tuning | 574 779 ± 22 280 | 59.80 ± 7.40% |
| X3 = V5 − multistage | 574 150 ± 29 315 | 59.80 ± 10.14% |
| X4 = V5 − expectimax search | 412 785 ± 2208 | 30.16 ± 0.38% |

V0–V5 present the incremental improvements to the best design (V5).
X1–X4 present the ablation studies of the best design.

score to 625 377, and the 32768-tile reaching rate to 72% with 6-ply search. Interestingly, this technique also significantly improved the chance of creating 65536-tiles to roughly 0.02% with 3-ply search, which is the highest 65536-tile reaching rate. Note that 65536-tiles were not reached for 5-ply and 6-ply searches since insufficient episodes (100 only) were evaluated for 5-ply and 6-ply.

*D. Comparison to SOTA*

In order to compare with the previous SOTA [5], [39], we used the same network, namely the 5×6-tuple network (Fig. 2) with 2-stage OTC with downgrading, and additionally we used a larger network, the 8×6-tuple network (Fig. 3) with 2-stage OTD+TC with downgrading. We tested 1-ply, 3-ply, and 5-ply search for these networks, and the results of the best $n$-tuple network of each method are shown in Table X. From Table X, both our methods outperformed the 16-stage 5×6-tuple network with TC in [5] in terms of fixed-depth search.

Furthermore, for the 2-stage 8×6-tuple network, we tested an additional 6-ply search with 100 episodes. The average search speed was about 2.5 moves/s by using a single thread of an Intel Core i9-7960X. The 2-stage OTD+TC 8×6-tuple network with 6-ply tile-downgrading expectimax search achieved an average score of 625 377, and a rate of 72% to reach 32768-tiles, which is also superior to those with the 1000ms search in [5]. Table IX shows the improvements of each technique from the non-optimistic method (V0) to the best design (V5) in the upper six lines, and ablation studies (X1 to X4) in the lower four lines. To our knowledge, the results are SOTA in terms of average score and the reaching rates of large tiles. Although we used three more 6-tuples, we only required 2-stage when compared to their work with 16-stage. Since each extra stage requires one more set of tuples, we use far less memory in total, namely 16 (2×8) 6-tuples, versus 80 (16×5) 6-tuples in [5].

However, compared with non-optimistic methods, our OI methods require more computing resources to explore large state spaces. The previous SOTA result [5] was trained with $4 \times 10^{10}$ actions, which corresponds to approximately 40M episodes or less. In contrast, our SOTA result was trained with 200M episodes. Nevertheless, such an amount of training is still affordable. It took us 66 hours to complete when using a C++ implementation [40] on an Intel E5-2698 v4 processor.

## V. Discussion

In this section, we discuss some other techniques related to 2048, including weight promotion [5], [26], redundant encoding [5], and carousel shaping [5].

Weight promotion can be used to speed up the multistage training by allowing the weights of $n$-tuple networks to be copied to the corresponding weights in the next stage upon its first access. This method is similar to OI in the sense that many weights are initialized higher than zeros. However, when compared to OI, their method only copies weights with small tiles, and those weights with large tiles, say 16384-tile in the next stage, are still initialized to zeros. In OI, these weights are large numbers leading to more exploration.

Redundant encoding improves the $n$-tuple network by adding some sub-tuples. OI may be applicable to this technique with an initialization scheme different from that in (12). Carousel shaping ensures each stage receives the same training amount. Since this paper focuses on the impact of OI, further research on incorporating these methods into OI is left open.

## VI. Conclusion

For the issue of exploration on 2048, we propose optimistic TD learning to improve the performance. Our approach significantly improves the learning quality of $n$-tuple networks. The significance of this paper is summarized as follows.

First, we improve the common TD methods with OI by using a hyperparameter $V_{init}$ to initialize the network weights, and demonstrate that OTD and OTC learning with $V_{init}$ = 320k significantly improve the performance, especially the chance of obtaining 32768-tiles, as shown in Section IV-A. Second, we propose OTD+TC learning, a hybrid learning paradigm that combines the advantages of both TD and TC for encouraging exploration and exploitation, respectively. A hyperparameter $P_{TC}$ is used to control the proportion of TC fine-tuning phase. We observe that OTD+TC with $V_{init}$ = 320k and $P_{TC}$ = 10% outperforms both OTD and OTC learning for larger networks, as described in Section IV-B. Third, we show that we do not need as many stages as in [5], thereby significantly reducing the required network weights, since OI effectively improves the learning quality as shown in Section IV-C2.

Furthermore, we use tile-downgrading to improve the search quality as shown in Section IV-C3, and the results in Section IV-D shown to outperform those in the previous SOTA [5] in terms of average score and 32768-tile reaching rate. Namely, the design with a 2-stage 8×6-tuple network trained by OTD+TC with $V_{init}$ = 320k and $P_{TC}$ = 10% achieved an average score of 625 377 and a rate of 72% reaching 32768-tiles through 6-ply tile-downgrading expectimax search. In addition, for sufficiently large tests, 65536-tiles are reached at a rate of 0.02%.

We conclude that for 2048 and similar games, learning with explicit exploration is useful even if the environment seems to be stochastic enough. This paper demonstrates that the optimistic method is promising for stochastic games.

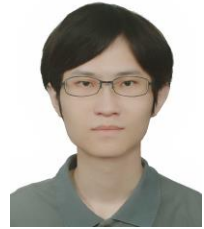

**Hung Guei** is currently a Ph.D. candidate in the Department of Computer Science at National Yang Ming Chiao Tung University. His research interests include artificial intelligence, machine learning, computer games, and grid computing.

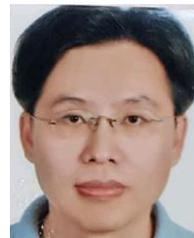

**Lung-Pin Chen** (M'15) received the M.S. degree from the National Chung-Cheng University, Chiayi, Taiwan, in 1993, and the Ph.D. degree from the National Yang Ming Chiao Tung University, Hsinchu, Taiwan, in 1999, all in computer science. He is an Associate Professor at the Department of Computer Science and Information Engineering, Tunghai University, Taichung, Taiwan. His research interests include distributed algorithm, cloud computing, computational intelligence, and machine learning.

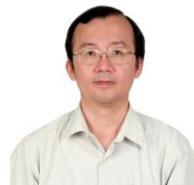

**I-Chen Wu** (M'05-SM'15) is currently the executive officer of Artificial Intelligence Computing Center at Academia Sinica, a research fellow of Research Center for IT Innovation at Academia Sinica, and also a professor of the Department of CS at National Yang Ming Chiao Tung University. He received his B.S. in Electronic Engineering from National Taiwan University (NTU), M.S. in Computer Science from NTU, and Ph.D. in Computer Science from Carnegie-Mellon University, in 1982, 1984 and 1993, respectively. He serves the editor-in-chief of ICGA Journal and an associate editor in the IEEE Transactions on Games. He currently serves as the vice president of the International Computer Games Association, and the president of the Taiwanese Computer Games Association; and served as the president of the Taiwanese Association for Artificial Intelligence in 2015-2017.

His research interests include computer games and deep reinforcement learning, and his research achievements include several state-of-the-art game playing programs, such as CGI for Go and Chimo for Chinese chess, winning over 30 gold medals in international tournaments, like Computer Olympiad. He wrote over 150 technical papers, and served as chairs and committee in over 30 academic conferences and organizations, including the conference chair of IEEE CIG conference 2015.